\documentclass{article}

\usepackage{microtype}
\usepackage{graphicx}
\usepackage{subcaption}
\usepackage{booktabs} 
\usepackage[table]{xcolor}
\usepackage{enumitem}

\usepackage{hyperref}

\usepackage[accepted]{icml_styles/icml2025}

\usepackage{amsmath}
\usepackage{amssymb}
\usepackage{mathtools}
\usepackage{amsthm}
\usepackage[customcolors]{hf-tikz}

\usepackage[capitalize,noabbrev]{cleveref}

\theoremstyle{plain}

\theoremstyle{definition}

\theoremstyle{remark}

\usepackage[textsize=tiny]{todonotes}

\usepackage{amsmath,amsfonts,bm}

\def\eqref#1{equation~\ref{#1}}

\def\1{\bm{1}}

\DeclareMathAlphabet{\mathsfit}{\encodingdefault}{\sfdefault}{m}{sl}
\SetMathAlphabet{\mathsfit}{bold}{\encodingdefault}{\sfdefault}{bx}{n}

\DeclareMathOperator*{\argmin}{arg\,min}

\usepackage{url}
\usepackage{parskip}
\usepackage{amsfonts}
\usepackage{multirow}
\usepackage[ruled,linesnumbered]{algorithm2e}

\icmltitlerunning{Can Memory-Augmented Language Models Generalize on Reasoning-in-a-Haystack Tasks?}

\begin{document}

\twocolumn[
\icmltitle{Can Memory-Augmented Language Models Generalize \\ on Reasoning-in-a-Haystack Tasks?}



\icmlsetsymbol{equal}{*}

\begin{icmlauthorlist}
\icmlauthor{Payel Das}{equal,yyy}
\icmlauthor{Ching-Yun Ko}{equal,yyy}
\icmlauthor{Sihui Dai}{equal,yyy,xxx}
\icmlauthor{Georgios Kollias}{yyy}
\icmlauthor{Subhajit Chaudhury}{yyy}
\icmlauthor{Aurelie Lozano}{yyy}
\end{icmlauthorlist}


\icmlaffiliation{yyy}{IBM AI Research}
\icmlaffiliation{xxx}{Princeton University (Work done during internship at IBM Research)}

\icmlcorrespondingauthor{Payel Das}{daspa@us.ibm.com}

\icmlkeywords{Machine Learning, ICML}

\vskip 0.3in
]



\printAffiliationsAndNotice{\icmlEqualContribution} 

\begin{abstract}
Large language models often expose their brittleness in reasoning tasks, especially while executing long chains of reasoning over context. We propose MemReasoner, a new and simple memory-augmented LLM architecture, in which the memory learns the relative order  of facts in  context, and enables hopping over them, while the decoder selectively attends to the memory. MemReasoner is trained end-to-end, with optional  supporting fact supervision of varying degrees. We train MemReasoner, along with   existing memory-augmented transformer models and a state-space model, on  two distinct synthetic multi-hop reasoning tasks. Experiments performed under  a variety of challenging scenarios, including the presence of long distractor text or target answer changes in test set,  show strong generalization of MemReasoner on both single- and two-hop tasks. This  generalization of MemReasoner  is achieved  using  none-to-weak supporting fact supervision (using none and  1\% of supporting facts for one- and two-hop tasks, respectively). In contrast, baseline models overall struggle to generalize and benefit far less from using full supporting fact supervision. The results highlight the importance of  explicit memory mechanisms, combined with additional weak supervision, for improving large language model's context processing ability toward reasoning tasks. 
\end{abstract}

\section{Introduction}
Transformer-based large language models (LLMs)  have recently shown impressive performance in many natural language processing (NLP) tasks, including machine translation, question answering, and reading comprehension, demonstrating the signature of general reasoning abilities. Despite these achievements, LLMs often fail to generate accurate information with respect to the context and are prone to hallucinations. Perhaps surprisingly, such hallucinations are found even on simple tasks that require some form of reasoning over the context. Oftentimes, irrespective of the specific type of reasoning involved, LLM hallucinations occur from failure to resolve a dependency chain over the input, rather than memorizing the exact training sequence. This issue of incorrectly modeling long-range dependency and reasoning has been reported for a variety of tasks including logical reasoning \cite{levy2024tasktokensimpactinput, kuratov2024babilong, wan2024logicaskerevaluatingimprovinglogical} and algorithmic reasoning \cite{liu2023transformerslearnshortcutsautomata, liu2023exposingattentionglitchesflipflop}.  

In this work, we provide a novel language model architecture that is designed to naturally handle iterative processing over the context to learn long-range dependencies. We refer to this model as MemReasoner, which is a memory-augmented language model enhanced with two basic operations: (i)  explicit learning of temporal orders of facts/events present within the context, and (ii) a mechanism for iteratively  reading from the context and updating the query accordingly. We further explore benefits of utilizing none--to--weak supporting fact supervision (along with final answer supervision) during model training.

Multiple  synthetic benchmarks \cite{hsieh2024rulerwhatsrealcontext, kuratov2024babilong, liu2023exposingattentionglitchesflipflop} have been recently proposed to stress test different language models, which are designed to isolate and probe simple reasoning errors, such as temporal awareness, coreference resolution, fact chaining. These synthetic benchmarks complement well the real-world benchmarks like Long Range Arena \cite{tay2020long} and BigBench \cite{srivastava2023beyond} and have been successfully demonstrating existing ``reasoning'' gap in current language models \cite{hosseini2024llmreasonerscreatedequal}. We stress test MemReasoner, along with a number of existing baseline transformer-based and alternative language models, on two distinct synthetic tasks that require performing multiple hops over the input to   track  or bridge  over entities.  We evaluate how the task-finetuned models generalize to different test scenarios, which  involve finding unseen response from the context, resolving the dependency chain over longer context that contains unseen hard or soft distractor text, or a combination of both. We also experiment with a  generalization scenario, where a model trained on the two-hop task is tested on a single-hop version  but includes longer instances.    


Our main contributions are:
\begin{itemize}[leftmargin=*, noitemsep]
    \item  A novel memory-augmented LLM architecture, namely MemReasoner, which is trained to execute \textit{temporal processing and iterative read} over the context written to a latent memory module that is separate from the transformer decoder. 
    For this purpose, a  positional encoding  scheme is proposed to learn the relative order of facts in memory, that helps with  selective attention of the transformer decoder to the memory. 
    \item We evaluate the proposed architecture, along with a number of language model baselines with and without a  memory, on \textit{two synthetic multi-hop reasoning-in-a-haystack tasks} (1) babi - that include 1-hop and 2-hop logical reasoning tasks and (2) variable tracking (VT)- that require tracking entities with 1- and 2-hop connections. 
    \item We subject the task-finetuned models to a \textit{variety of challenging generalization scenarios}. 
    \item Results show that MemReasoner, when compared to existing recurrent baselines, including a memory-augmented recurrent transformer (RMT) model~\citep{bulatov2022recurrent} and Mamba~\citep{gu2023mamba}, a state-space model,  \textit{generalize better} in the single hop task setting, where target answer, context length with hard/soft distractors, task complexity, or a combination of thereof, differs from training to test distribution.
    \item On tasks that require two hops, MemReasoner \textit{benefits from additional weak supervision on supporting facts}.  With usage of  supporting facts only on 1\% of training samples,  the proposed model better generalizes to many different test scenarios, when compared to the baselines that utilize 100\% of available supporting fact samples. 
\end{itemize}
\section{Related Work}

\textbf{Modeling long-range dependency}
Many tasks require multiple steps to be executed effectively and in the right order, which include, for example,  logical and mathematical reasoning  and multistep knowledge editing.  Current transformer-based LLMs are known to produce erroneous output in those scenarios, possible reasons include learning shortcuts from training data bias \citep{ju2024investigatingmultihopfactualshortcuts,ruder2021challenges,mitchell2023we, wu2024cofcastepwisecounterfactualmultihop, levy2024tasktokensimpactinput}, fragile internal mechanisms  like attention glitches \citep{liu2023exposingattentionglitchesflipflop}, and attention sink \citep{xiao2024efficientstreaminglanguagemodels,liu2023lost}, which can be attributed to lack of recurrence in self-attention.  At a high level, such failures can be attributed to a lack of ``System-2'' like thinking mode \citep{kahneman2011thinking}, which encourages deliberative and logical thinking steps, in vanilla transformer language models.    

To address this limitation, recent works have attempted to include few-shot 
\citep{brown2020language, min2022rethinkingroledemonstrationsmakes}
and chain of thought (CoT) prompting 
\citep{wei2022chain},  providing access to external tools/reward models/verifiers \citep{schick2023toolformerlanguagemodelsteach,khalifa2023gracediscriminatorguidedchainofthoughtreasoning}, etc.  Augmenting language models with memory modules has been proposed, 
e.g.,  in \citep{nye2021workscratchpadsintermediatecomputation} the model is asked to output immediate reasoning steps to a “scratchpad” which is then recurrently processed by the model. Another promising research direction  is to train the transformer model with an external latent memory module \citep{das2024larimarlargelanguagemodels} or with additional learnable memory tokens \citep{burtsev2021memorytransformer}. Later, architectures that include segment-level  recurrent  processing over internal memory tokens  have been proposed, e.g.,  Transformer-XL \citep{dai2019transformerxlattentivelanguagemodels} and Recurrent Memory Transformer (RMT) \citep{bulatov2022recurrent}. This line of work has shown the ability to process very long input 
and has emerged as a promising path for modeling long-term dependencies and exploiting memory processing ability for tasks like algorithmic and reasoning. 


Structured state space models such as Mamba \citep{gu2023mamba} have recently emerged as a promising alternative to self-attention layers and transformers for sequence modeling. A differential feature of Mamba is the selection mechanism, i.e., the context-aware ability to focus on or filter out inputs into a fixed-size sequential state. Mamba offers faster inference due to its fixed-memory recurrent architecture, which allows for efficient processing of long sequences. However, this constant-memory also can make the in-context recall ability brittle, compared to transformers \citep{jelassi2024repeat,waleffe2024empiricalstudymambabasedlanguage,park2024mambalearnlearncomparative}. 


In MemReasoner, the segment-level processing takes place in the latent memory module, which is further augmented with a recency awareness and iterative read mechanism. This is    inspired by the distinction between System 1 and System 2-like thinking \citep{kahneman2011thinking}, MemReasoner utilizes the decoder for fast output generation and the memory module for slow processing of the input, which are the two components tightly integrated via training. Generation of intermediate steps in MemReasoner is analogous to CoT method. And,  using optional weak supervision on supporting facts in MemReasoner is similar to the line of works that uses rationales for supervised fine-tuning or for preference tuning of LLMs to enhance their reasoning abilities \citep{zelikman2022starbootstrappingreasoningreasoning, pang2024iterativereasoningpreferenceoptimization}.  

Different from recurrent passing of the global memory tokens from the previous segment to the next segment within the transformer layers themselves, as done in Transformer-XL and RMT,  MemReasoner performs multiple hops over the ``ordered'' segment encodings stored in memory, updates the query accordingly, and provides only the final readout(s) to the decoder.  In contrast to ``scratchpad'' and CoT line of works, MemReasoner does not maintain a memory of  explicit (generated) tokens, rather operates over the latent  encodings of context stored in memory. 

\textbf{Long-range dependency and reasoning benchmarks}
Many datasets and benchmarks have been designed to isolate issues with long-range dependency modeling and generalization. Synthetic tasks have become increasingly popular for testing language models, as those, compared to real-world language datasets, provide a cleaner and more controlled setup for probing the abilities and limitations of transformer-based language models, which is why we evaluate MemReasoner and baselines on two synthetic tasks here (see section \ref{sec:data_preprocessing} for details). A number of recent benchmarks include synthetic tasks that require multi-hop processing, e.g., \citep{hsieh2024rulerwhatsrealcontext,kuratov2024babilong,li2024needlebenchllmsretrievalreasoning, liu2023exposingattentionglitchesflipflop}. One prevalent direction covered in those benchmarks is generalization to long input length, where long irrelevant (hard) distractor text has been added to the original input to artificially lengthen it. While retrieving a ``needle'' (a piece of text) from the  long context input (``haystack'') has been focus of many of the synthetic tasks, more recently,  benchmarks have been proposed, which include tasks that require learning long-term dependencies over multi-hop connections, see Figure \ref{fig:example} for example. These benchmarks have become a natural testing ground for many long-context transformer-based LLMs as well as alternative architectures like state-space models \citep{gu2023mamba}. While some experimental settings consider  training models, to be  tested on those ``reasoning''-in-a-haystack tasks, on  sequences that include the irrelevant distractors of different lengths, here we focus on generalization from training samples that do not include the distractor text. 

We also extend testing model's generalization beyond using long inputs with hard distractors in the following ways: 
(i) We create a new dataset that contains long input sequences that contain soft distractors. 
(ii) We include checking generalization to a distribution that contains test samples in which target answer is different from what is seen during training, which can be further combined with input length generalization. (iii) We also consider varying task complexity,  i.e., a model trained on a two-hop task is tested on corresponding single-hop task, which again can include longer test sequences with distractors. 

\section{Multi-step Reasoning with MemReasoner}
\begin{figure*}[t!]
    \centering
    \includegraphics[width=1\linewidth]{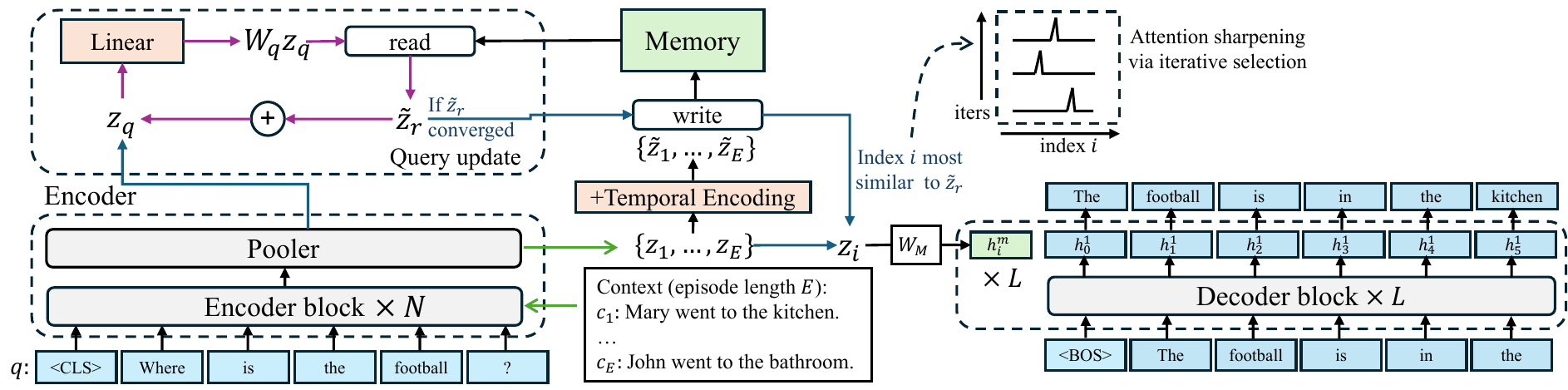}
    \caption{A diagram of the pipeline for reasoning with MemReasoner. \textcolor{black}{(a) Conceptual overview of the framework. (b) Detailed architecture.}  $q$ denotes the query, $c_1, ..., c_E$ denotes the context for answering the query.  $z_q$ denotes the encoding of the query while $\{z_1, ..., z_E\}$ denote encodings of each line of the context.  We use $\tilde{z}$ to denote temporally encoded latents.}
    \label{fig:reasoning_pipeline}
\end{figure*}
The key components of MemReasoner involve an LM encoder, an episodic memory module, and an LM decoder (see  Figure \ref{fig:reasoning_pipeline}a). The role of the episodic memory module is to enable \textit{write} of the context encodings in the memory,  to allow performing search over the context encodings and \textit{read} from them, in order to feed the decoder to execute the task. 
Given a reasoning task, for which the final answer is available, the MemReasoner architecture is trained to recover 
the final answer,  with or without using the groundtruth supporting fact supervision. An additional point worth mentioning is that, MemReasoner is also trained to learn the relative order of the facts in the context and use that for memory write/read, which can be crucial for explicitly maintaining temporal ordering of context.
A search in the latent memory space is performed during training, to correctly output the final answer (and optionally the supporting facts). This imposes a ``selection'' mechanism over the memory of the context and helps ``sharpening'' decoder's attention over the fact that matters for correct output generation.    
Note that, the use of supporting fact supervision is optional: we consider training scenarios involving none-to-full supporting fact supervision with different loss objectives to study the benefits of various degrees of supervision.
Details are provided below.

\subsection{Using the Larimar Framework as Backbone for MemReasoner}
\label{subsec:larimar}

The backbone memory-augmented LLM used in this study is Larimar~\citep{das2024larimarlargelanguagemodels} (although our approach could in principle be used in conjunction with other LLMs augmented with an additional (episodic) memory module).
Let $\mathcal{X}$ be the LM input space, $\mathcal{Z}$ be the latent space, and $\mathcal{Y}$ be the LM output space. Larimar features an encoder $e$ that maps an input to an embedding $z\in \mathcal{Z}\subseteq \mathbb{R}^{D}$, and a memory module $\mathcal{M}$. The memory $\mathcal{M}$ is adaptable in the sense, that  it allows ``write'' and ``read'' operations as episodes (aka, contexts $C$, where each context is comprised of $E$ sentences) arrive, i.e., $\hat{M}=write(M,z), z_{\text{read}}=read(\hat{M},z)$, wherein $\hat{M}$ is the updated memory after an write. And, a decoder $d$ that performs generations conditioned on the memory readout $z_{\text{read}}$. 

Now, suppose one is given an input context $C=\{c_1, ..., c_E\}$ with $E$ denoting the length of the context, and the target task is to answer a question $q$ conditioned on the given context $C$. To approach the task within the Larimar framework,  the input, both context $C$ and query $q$, are encoded to their latents ($z_1,\ldots,z_E$ and $z_q$) via the encoder $e$. Next, let $M_0$ be the initial memory,  write the context to the memory via a $write$ operation. \textcolor{black}{To do so, Larimar follows the earlier works on Kanerva Machine \citep{wu2018kanerva}, 
where the memory is viewed as a global latent variable in a generative model. In this framework, the goal is to learn a memory dependent data prior and learnable addresses, where the memory update and read/write are considered as Bayesian inference, i.e., the posterior parameters are updated as new data arrives.  \citep{pham2022generative} reformulated the Bayesian updates for encoding new memories and for decoding data from memories into an equivalent minimization problem, which essentially amounts to solving a linear system of equations and can be efficiently done via computing matrix pseudo inverses indicated by $\dagger$ hereafter.} 

Specifically, memory is updated via the \textit{write} operation such that, $\hat{M}=(Z_\xi M_0^\dagger)^\dagger Z_\xi$, where $Z_\xi=[z_1+\xi_1,z_2+\xi_2,\ldots,z_E+\xi_E] $ and $\xi_i\sim \mathcal{N}(0,\sigma_\xi^2 I)$. 
Then, the $read$ operation translates the query embedding from the lens of the encoded memory to a query readout $z_r$ via $z_r=(z_q\hat{M}^\dagger+\eta)\hat{M}$, where $\eta\sim\mathcal{N}(0,\sigma_\eta^2I)$. Lastly, the decoder $d$ decodes the query $q$ conditioned on the readout by using a \textit{learnable} broadcasting matrix of parameters $W_M$ that casts $z_r$ to each decoder layer and obtains $h_k^m$ that serves as the past key values for $k=1,\ldots,L$, where $L$ is the number of layers in the decoder. 

We use the memory-augmented LLM architecture of Larimar and the above operations as the backbone for MemReasoner, due to its memory and space-efficient read/write abilities and demonstrated generalizability at test-time. 

However, as such, Larimar's mechanisms are insufficient to handle complex tasks, such as making multi-hop connections over the context.  We now introduce key additional features provided in MemReasoner to handle such tasks: (i)  explicit learning of temporal orders of facts/events present within the context, and (ii) mechanism for iteratively  reading from the context and updating the query accordingly.

\subsection{Memory with Temporal Order}
\label{subsubsec:temporal}

In Larimar, the latent encoding of facts $\{z_1, ..., z_E\}$ within a context episode $C$ are written in the memory $M$ in an order-invariant manner. However, many multi-step reasoning tasks require some notion of temporal context. For example, when answering ``where is John?'' in the context of ``$[\ldots]$ John is in the bathroom. $[\ldots]$ John goes to the garden.'' (`$[\ldots]$' denotes irrelevant facts), there should be a mechanism in place to guarantee that the memory encodes the correct temporal order of the facts, and the readout  should reflect ``John goes to the garden.'' as the supporting fact instead of ``John is in the bathroom.''. 

To introduce some temporal notion within the context (i.e. facts that appeared later in context are more recent), in MemReasoner we introduce a temporal encoding module $\mathcal{P}$ that transforms \textit{un-ordered} fact latents $\{z_1, ..., z_E\}$ within a context episode to their \textit{ordered} counterparts $\{\tilde{z}_1, ..., \tilde{z}_E\}$. 
In this paper, we will mainly leverage a learnable bidirectional GRU for temporal encoding, i.e., 
\begin{align*}
    \{\tilde{z}_1, ..., \tilde{z}_E\}\leftarrow \text{GRU}(\{z_1, ..., z_E\})
\end{align*}
This temporal encoding module $\mathcal{P}$, however, is generic and allows any structure featuring sequentiality within contexts. Additional discussion about this design choice is provided in the Appendix \ref{app:temporal}.
These ordered context embeddings $\{\tilde{z}_1, ..., \tilde{z}_E\}$ are then written to memory via Larimar's $write$ operation.  

It is worth mentioning that earlier works on memory-augmented neural nets, which use a recurrent neural net together with an external memory, have investigated ideas like temporal feature learning and iterative hops over context, for example, see  \citep{weston2014memory, sukhbaatar2015endtoendmemorynetworks}. However, to our knowledge, this is the first study to enable those operations around a latent memory coupled to a transformer-based LLM during training and to test the resulting model's generalizability  on  long-range dependency learning.

\subsection{Iterative Read And Query Update}
\label{subsubsec:qupdates}
 A typical multi-step reasoning task often inherently requires ``hops'' between facts until the final solution is found. Additionally, the query embedding can be updated accordingly to reflect the most recent hop.

 In order to perform hopping between facts, we first recall the three key components interacting with the memory module $\mathcal{M}$, the fact embeddings ($\{z_1,\ldots,z_E\}$) within a context episode, the query embedding $z_q$, and the memory readout $z_r$. Let us further 
 assume that $\mathcal{M}$ 
 stores facts that have  been ordered temporally $\{\tilde{z}_1, ..., \tilde{z}_E\}$. 

To enable \textit{iterative read}, we pass $z_q$ through a linear layer to obtain $\hat{z_{q}}$=$W_qz_q$ before the $read$ operation from the memory, where $W_q \in \mathbb{R}^{D\times D}$ is a learnable parameter that absorbs the scale changes introduced by the position encoding in the memory. Specifically, different from Section~\ref{subsec:larimar}, here we have $z_r=(\hat{z_{q}}\hat{M}^\dagger+\eta)\hat{M}$.

To \textit{update the query}, we first update the query latent and let $z_q\leftarrow z_q + \alpha\cdot z_r$, where $\alpha \in \mathbb{R}$ is a hyperparameter to balance the load from the previous readout. The updated query is then fed into the memory module for another $read$ operation to obtain a new $\tilde{z}_r$. 
The query update procedure is repeated until the readout converges (i.e. $||\tilde{z}^t_r - \tilde{z}^{t+1}_r||_2 < \tau$ where $\tilde{z}^t_r$ denotes the readout at time $t$ and $\tau$ is a hyperparameter) or until it reaches a fixed number of maximum iterations.
 
\subsection{Full Workflow}
Now that we have discussed all components of MemReasoner, we elaborate  the full pipeline in the following and provide a visualization  in Figure \ref{fig:reasoning_pipeline}.

Consider an input context $C=\{c_1, ..., c_E\}$, a question $q$, an encoder $e$, a temporal encoding module $\mathcal{P}$, an initial memory module $\mathcal{M}$, and a decoder $d$. We first encode the context $C$ and query $q$ to their latents, $z_1,\ldots,z_E$ and $z_q$, via encoder $e$. Then, we follow Section~\ref{subsubsec:temporal} and transform $z_1,\ldots,z_E$ to $\tilde{z}_1, ..., \tilde{z}_E$. Next, we write the ordered context $\tilde{z}_1, ..., \tilde{z}_E$ to the memory and obtain $\hat{M}$. Subsequently, we read using the query latent from the memory $\hat{M}$ and perform query and read updates according to Section~\ref{subsubsec:qupdates}. 
After we have obtained a $\tilde{z}_r$ as a final readout which does not undergo update anymore, we map $\tilde{z}_r$ to the corresponding unordered encoding in $M$. This is because we only want the additional position information to be used when locating the most relevant contexts, but not during the decoding - if being fed to the decoder, the decoder may overfit to the ordering information in the latents. We do this by first finding the index of the most similar ordered latent encoding $i = \argmin_{j \in \{1,...,E\}} ||\tilde{z}_r - \tilde{z}_j||_2$ and then obtaining the corresponding encoding $z_i$ from the unordered encodings  (prior to undergoing temporal encoding in Figure \ref{fig:reasoning_pipeline}) $\{z_1 ... z_E\}$.
Lastly, the decoder $d$ decodes the prompt $P_a$ given for answer generation conditioned on $z_i$. 
We provide the full pseudocode in Algorithm \ref{alg:inference} in the Appendix \ref{app:algorithm_description}. 

\subsection{Inference-Time Update (IU)}
To generalize on long-context tasks with a memory-augmented LLM, a potential hurdle can come from the two fundamental operations of the memory, write and read, as described in Section~\ref{subsec:larimar}. In both operations, numerical solves of the linear systems involve computing matrix pseudo inverses, which can be unstable when the matrix has many more columns than rows or the other way around. Secondly, encoding the temporal order in very long facts with a GRU can further incur vanishing or exploding gradient. 

To cope with these, we further introduce an optional inference-time update (IU) step, where we dynamically filter irrelevant contexts before re-encoding the temporal order and the memory operations. Specifically, with the first pass of memory write and query read, we identify a subset of contexts that are most relevant to the query by their proximity in the ordered latent space ($||\tilde{z}_r - \tilde{z}_j||_2$). Then, we re-encode the temporal order on this remaining contexts (much shorter) to get their ordered counterparts for the subsequent memory write and query read.
This optional update step enables the model to focus on the most relevant information  and avoids numerical instabilities. All MemReasoner performance values in the main paper are obtained using IU, unless otherwise stated. 

\subsection{Training Objectives}
Let $\mathcal{D}_{\text{pretrain}}$ denote the pretraining data distribution, while $\mathcal{D}_{\text{finetune}}$ denotes the data distribution corresponding to the reasoning task. Each sample from $\mathcal{D}_{\text{finetune}}$  is of the form $(q, C, a)$ where $q$ is the query, $C = \{c_1, ..., c_E\}$ are the facts in the context, and $a$ is the answer. Depending on the availability of supporting fact information in each sample, we let $\mathcal{D}_{\text{finetune}}^*$ be the subset of samples with groundtruth supporting fact, i.e. each sample from $\mathcal{D}_{\text{finetune}}^*$ is of the form $(q, C, a, S)$, where $S$ is a set of indices corresponding to the supporting facts (we will use $S_i$ to denote the $i$th supporting fact index in $S$). Meanwhile, the pretraining distribution corresponds to a generic corpus, e.g. Wikipedia. Remember, $e$ denotes the encoder, $d$ denotes the decoder, $t$ denotes temporal encoding, $\tilde{z}_r^i$ denotes the $i$th temporally encoded readout from iterative reading with $\tilde{z}_r^0 = q$, $z_r^i$ represent the unordered encoding corresponding to the $i$th ordered readout, and $P_a$ denotes the prompts for generating the answer.  To train the model, we utilize the following loss function.

\tikzset{
h1/.style={
set fill color=blue!10,
set border color=blue!10,
},
h2/.style={
set fill color=red!10,
set border color=red!10,
}
}

{\small
\begin{equation}
\label{train_loss}
\begin{split}
 &L = \rho \underbrace{\mathbb{E}_{x \sim \mathcal{D}_{\text{pretrain}}} \ln p(d(e(x)))}_{\text{autoencoding of pretraining dataset}}\nonumber \\ 
& +\mathbb{E}_{(q, C, S, a) \sim \mathcal{D}_{\text{finetune}}}\underbrace{\mathbb{E}_{z_{r}^{|S|} \sim p(z_{r}^{|S|} | q, M, \tilde{z}_{r}^{0} ... \tilde{z}_{r}^{|S|-1})} \ln p(a | z_{r}^{|S|}, P_a)}_{\text{reconstruction of answer}
}  \nonumber\\
&\tikzmarkin[h2]{b}(6.55,3.55)(-0.2,-0.9)
+\mathbb{E}_{(q, C, S, a) \sim \mathcal{D}_{\text{finetune}}^*} \left[ \delta \underbrace{\sum_{i=1}^{|S|} \mathbb{E}_{\tilde{z}_{r}^{i} \sim p(\tilde{z}_{r}^{i} |q, M, \tilde{z}_{r}^{0} ... \tilde{z}_{r}^{i-1})} \ell_{\text{order}}(\tilde{z}_{r}^{i}, S_i)}_{\text{ordering loss}} \right]\tikzmarkend{b}\\
\end{split}
\end{equation}
}
where $\delta$ and $\rho$ are hyperparameters controlling regularization strength and $\ell_{\text{order}}$ is given by
{\small
\begin{equation*}
\begin{gathered}
    v(z_r) = \text{softmax}([-||t(e(c_1)) - z_r||_2, ..., -||t(e(c_E)) - z_r||_2]^\intercal) \\
    \ell_{\text{order}}(z_r, s) = -\ln v(z_r)_s
\end{gathered}
\end{equation*}
}


In the overall loss $L$, the first term corresponds to the autoencoding loss on the pretraining dataset. The second term is the reconstruction loss of the answer with respect to the corresponding prompt for obtaining the answer $P_a$ and final readout. 
Depending on the degree of supporting fact supervision, we further add an optional third loss encouraging the index of the most similar entry (by l2 distance) to the ordered readout at each iteration to match the index of the supporting fact through computing the cross entropy. We note that $\mathcal{D}_{\text{finetune}}^*$ can be a smaller, as little as 1\% of the training set size, subset of $\mathcal{D}_{\text{finetune}}$, covering the realistic scenario  when supporting fact supervision is scarce.


\section{Experimental Details and Results}






\subsection{Task Datasets and Data Pre-processing}
\label{sec:data_preprocessing}
We utilize two sets of tasks, both of which involve fact chaining by drawing  multi-hop connections over the context. Namely,  
we resort to  tasks 1 and 2  from  the synthetic bAbi benchmark \citep{weston2015towards}, 
and the Variable Tracking (VT) task from the RULER benchmark~\citep{hsieh2024rulerwhatsrealcontext}. 

\textbf{bAbi Tasks.} The bAbi datasets were prepared by synthesizing relations among characters and objects across various locations, each represented as a fact, such as ``Mary traveled to the garden". Task 1 requires performing a single hop to find the answer, whereas task 2 requires gathering two supporting facts in the right order (see Fig \ref{fig:example}). For preprocessing bAbi data, we treat each training sample comprised of multiple facts as a single context episode, and individual sentence within that context as an instance within that episode. Each fact within an episode contains up to 64 tokens. In our experiments, we consider training models with bAbi task 1 and task 2 data.  To evaluate the generalization of these models, we consider harder variants of bAbi:

\textit{BABILong. } BABILong extends these single and multi-hop tasks to a long context setting \citep{kuratov2024babilong}
where the difficulty of the task is further varied by changing the length of the irrelevant (PG-19 distributed) text added to the initial bAbi samples.  
The BABILong leaderboard shows  tasks 1 and 2, while simple, are  challenging enough for off-the-shelf LLMs to solve in the presence of hard (irrelevant) distractor text. 
For BABILong (used during inference), if sentences are longer than 64 tokens, we  split the sentences at multiples of 64 tokens. 

\textit{BABILong-soft. } We create a variant of BABILong that contains soft distractors (in-distribution padding) instead of hard ones in the original dataset, which we refer to as BABILong-soft. Using this padding, we increase context length up to 4k tokens.  We provide details about how this data is generated in \cref{app:dataset_details}.

\textit{Location change. } We also evaluate the robustness of models to modified bAbi test samples that have locations that are unseen during training. 
Specifically,  we map locations mentioned in bAbi samples to a different location (ie. office $\to$ library). We provide details about all location mappings in \cref{app:dataset_details}. 

\textbf{Variable Tracking (VT) tasks.} Variable tracking, introduced by RULER benchmark \citep{hsieh2024rulerwhatsrealcontext}, emulates a minimal coreference chain resolution task, which requires tracking relevant co-occurrence patterns and inferring skipped connections to bridge linked entities over long context.  Specifically, the model is given context with lines with information about variable value assignment such as ``VAR AAAAA = 16438'' or ``VAR BBBBB = AAAAA'', and the model is prompted to obtain all variables with a specific value. Variable names have the format of 5 repeating letters randomly sampled from the alphabet.  The task complexity is further increased by adding  more chains. We provide an example and additional data details in \cref{app:var_track_gen}.

We train and evaluate with chains of length 2, 4, 6, 8, and 10 and return the average accuracy over all chain lengths for the 1-hop and 2-hop VT tasks.  In order to pad the context for lengths 1k, 4k, and 16k during test, we follow the approach taken from RULER of padding with the following sentences as hard distractors ``The grass is green. The sky is blue. The sun is yellow. Here we go. There and back again.$\backslash$n'' until the context reaches the desired length.  This noise is not present during training and the 0k data follow the same distribution as the training data.

\subsection{Baseline Models}
\label{sec:baselines}
We  benchmark  RMT and  Mamba models that are fine-tuned on individual task  and test them  on corresponding bAbi/BABILong  and VT test samples. 
Specifically, we fine-tune off-the-shelf RMT and Mamba models using the next token prediction loss on final answer reconstruction and optionally on supporting fact (SF) reconstruction (100\% SF refers to all supporting facts used in training) on each task separately \footnote{We note that this reconstruction loss differs from the ordering loss used in training MemReasoner (Equation \ref{train_loss}) and instead matches a term we investigate in more depth in Appendix \ref{app:support}.  Training RMT or Mamba with the ordering loss is non-trivial and is beyond the scope of current work.}. To our knowledge, this study is the first one to explore reasoning generalization of RMT and Mamba that use SF supervision. We also evaluate an RMT-.14B model finetuned on shorter BABILong samples from \citep{kuratov2024babilong} on newly proposed BABILong-soft set. If not explicitly mentioned, size of the RMT model is .77B and the same for Mamba is 1.4B.  
We report task accuracy as the performance metric, so higher the better. 

Due to space constraints, we defer the training details~\ref{sec:training},  ablation studies \ref{app:ablation}, more baselines \ref{app:baselines}, and  additional experiments on supporting fact supervision \ref{app:support} in Appendix.  

\subsection{Results}
\label{sec:results}

We now present results on evaluating the generalization capabilities of MemReasoner and baseline models.  Across evaluation datasets, our results demonstrate that \textit{the design of MemReasoner allows it to take advantage of none--to-little supporting fact information, which largely improves its generalization capabilities.}  In comparison, other baselines generally struggle to generalize and do not benefit much from supporting fact information.

\begin{figure*}[th]
\centering
\includegraphics[width=\textwidth]{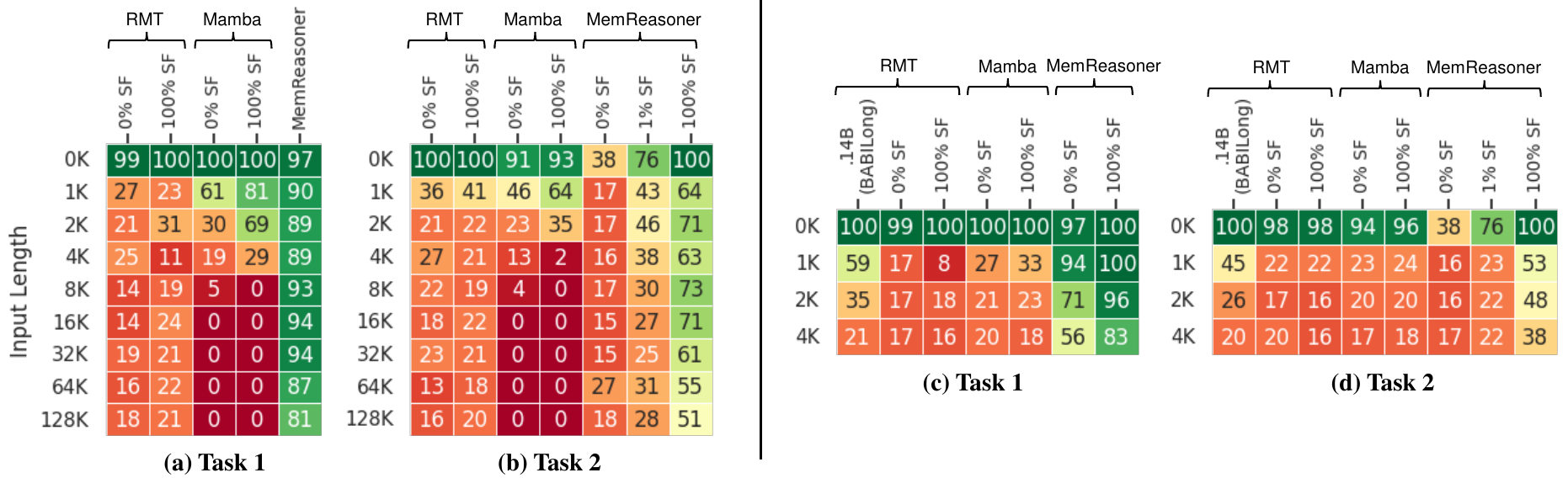}
  \caption{Performance of MemReasoner and baselines on varying lengths of BABILong (left) and BABILong-soft (right) data.}\label{fig:babilong_all}
\end{figure*}

\subsubsection{Generalization on long-context reasoning-in-a-haystack benchmarks}

\textbf{bAbi Tasks - BABILong Test.} Figure \ref{fig:babilong_all} (left) reports the accuracy of MemReasoner, together with baseline models, on  task 1 and task 2 test samples of varying length from BABILong, respectively. If supporting fact (SF) supervision is used during fine-tuning, it is mentioned explicitly as + x\% SF, where x denotes amount of training samples that have corresponding supporting facts. If a model without SF supervision provides high accuracy, we do not show results with + SF. RMT and Mamba, when finetuned on bAbi, achieve near-perfect accuracy on both tasks on original bAbi test samples. In contrast, MemReasoner trained with supervision on final answer only, while performing comparably on  task 1, it falls short on task 2. This low test accuracy of MemReasoner is due to making frequent errors at the second hop on the challenging 2-hop task. It is noteworthy that even a powerful baseline like    GPT-3 (175 B parameters)  with few-shot  and chain-of-thought prompting \citep{yang2023coupling} that performs  well on task 1, does much worse on task 2 that requires learning temporal dependence  and performing multiple hops across facts to generate the final answer of object location (see Appendix Table \ref{tab:babi}). Thus, we add   supporting fact supervision (using ordering loss in eqn. \ref{train_loss}) on 1\% of the training samples. On 0k test samples, MemReasoner + 1\% SF shows an accuracy improvement from  38\% to 76\%, while for MemReasoner + 100\% SF the accuracy becomes near perfect. This result suggests that MemReasoner benefits from a small amount of supporting fact supervision on a two-hop task.  
 
 bAbi-finetuned RMT and Mamba models 
  show a significant accuracy drop on BABILong samples beyond 0k input length, though both show  near-perfect accuracy at 0k. Interestingly,  additional supervision on 100\% supporting facts during RMT or Mamba training (in the form of supporting fact reconstruction loss) did provide none--to--little performance boost. 
In contrast, MemReasoner trained on bAbi generalizes well on BABILong for task 1, providing an average accuracy of 91.6\% and 89\% on $\le$ 8k and $\ge$ 16k BABILong samples, respectively.  
On task 2, 
MemReasoner also shows a performance drop from 0k to 1k.  
MemReasoner + 1\% SF 
shows more robust performance on BABILong, as accuracy drops from 76\% to only 43\%, as input length increases from 0k to 1k. MemReasoner + 100\% SF variant maintains  50-70\% accuracy for longer inputs up to 128k. 

\textbf{bAbi Tasks - BABILong-soft Test.} 
This long-context setting is  more challenging. Therefore, in addition to the bAbi-finetuned models,  we also include the top-performing models from BABILong leaderboard, i.e., an RMT-.14B model  finetuned on BABILong samples of up to 16k tokens, for comparison. Results are shown in Figure \ref{fig:babilong_all} (right) for tasks 1 and 2. On BABILong-soft task 1, all models lose accuracy compared to BABILong task 1. However, for MemReasoner, the performance drop is lowest throughout these context lengths. For task 2,  all models exhibit even more striking performance loss as input length increases, while MemReasoner with additional supporting fact supervision shows 
more robust performance in the 1-4k range.  

\begin{figure}[ht]
    \centering
    \includegraphics[width=1\linewidth]{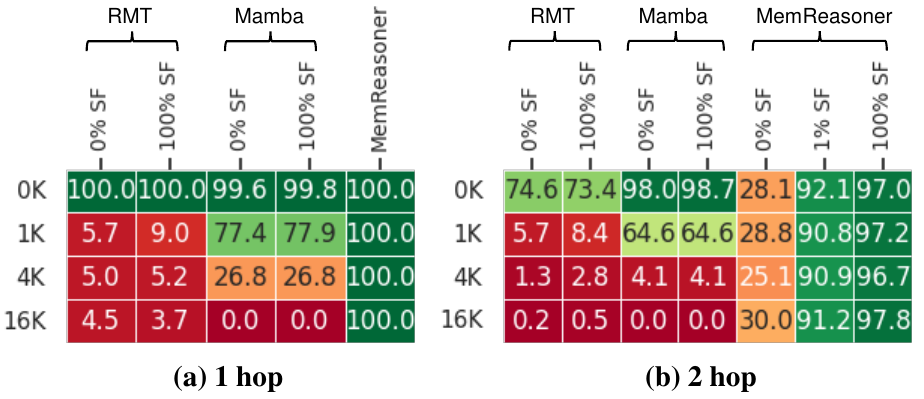}
    \caption{Performance on varying lengths of variable tracking.}
    \label{fig:VT}
\end{figure}

\textbf{Variable Tracking - Ruler Test.}
 Figure \ref{fig:VT} reports results on 1- and 2-hop variable tracking tasks. Remember that the longer test samples (1-16k length) from RULER benchmark do include additional noisy distractors in the input. 
In the single-hop VT experiments, MemReasoner maintains perfect accuracy over the entire context length range, whereas RMT shows a sharp drop from 100\% at 0k to 5.7\% at 1k. Even adding supervision on all  supporting facts does not help RMT's performance. Mamba performs  better than RMT at 1k, but also shows a stark accuracy lowering at 4k, which cannot be fixed with supporting fact supervision. 
 
For the 2-hop VT task, we observe that it is difficult to train RMT with 2 segments. RMT can easily learn a shortcut and have high accuracy on the training set, but does not generalize well to the test set at 0k length (74.6\% accuracy). At this length, we observe that RMT struggles with performance on longer chain lengths, achieving only 61.5\% accuracy on 8 chains and 3\% accuracy on 10 chains despite these chain lengths being present during training. Performance degrades further at longer context length. 
Again, access to supporting facts  appears ineffective in fixing the underlying issue. Mamba without and with supporting fact supervision shows consistently better performance than RMT. MemReasoner in the absence of supervision on intermediate steps does poorly at 0k (28.1\% accuracy), while maintaining it for longer inputs. With 1\% SF supervision, MemReasoner shows above 90\% accuracy for the entire 0-16k length range.


\subsubsection{Generalization to unseen target answers and from 2-hop → 1-hop}

 We create a new, more stringent testbed where the construct of the tasks remains the same, but the answer changes from training to test set (See section \ref{sec:data_preprocessing}). 
 As shown in Table~\ref{tab:location_swap},  when no SF supervision is used, on task 1 the accuracy order is RMT $<$ Mamba $<$ MemReasoner, while on task 2 the order is RMT $<$ MemReasoner $<$ Mamba. Given that, in this setting the length of the test sequence is similar to that of the training one, Mamba handles it better for task 2, with an accuracy of 34.5\%, though MemReasoner with 1\% SF supervision shows 64.9\% accuracy compared to 45.2\% for  Mamba with 100\% SF.

 \begin{table}[t]
    \centering
    \caption{Robustness to location changes in bAbi test set. Bold: highest; underlined; second highest.}
    \label{tab:location_swap}
    \resizebox{0.8\columnwidth}{!}{
    \begin{tabular}{c|cc}
    Model type & Task 1 & Task 2\\\hline
    \textcolor{black}{RMT}& 44.7 & 0.6\\
    \textcolor{black}{RMT + 100\%SF}& 31.3 & 0\\
    Mamba & \underline{74.4} & 34.5 \\
    Mamba  + 100\%SF & 52.7 & 45.2\\
    \hline
    MemReasoner  & \textbf{98.6} & 20.7\\
    MemReasoner  + 1\% SF & - & \underline{49.6}\\
    MemReasoner + 100\% SF & - & \textbf{63.9}\\
    \end{tabular}
    }
\end{table}

Finally, we also check if the models trained on 2-hop bAbi/VT can solve the simpler 1-hop version, but on the corresponding long context samples. We show their results in Tables \ref{tab:task2to1long} and \ref{tab:task2to1VT}. From the table, it can be seen that MemReasoner can generalize in this setting with no access to supporting facts for VT and with access to 1\% of supporting facts for bAbi. Baseline models show minimal generalization at longer input length for the single-hop task  and supporting fact supervision does not help much.  

\begin{table}[t]
    \centering
    \caption{Performance on bAbi task 2 → BABILong task 1 generalization.}
    \label{tab:task2to1long}
    \resizebox{0.8\columnwidth}{!}{
    \begin{tabular}{c|cccc}
    Model type & 0k & 1k & 2k & 4k\\\hline
    \textcolor{black}{RMT}& \textbf{100} & 19 & 20 & 12\\
    \textcolor{black}{RMT (bAbi) + 100\%SF}& 95 & 21 & 22 & 16\\
    Mamba& 86 & 33 & 24 & 7\\
    Mamba  + 100\%SF & \underline{99}& \textbf{55} & 27 & 9\\
    \hline
    MemReasoner & 45 & 12 & 22 & 14\\
    MemReasoner) + 1\% SF & 52 & \underline{45} & \textbf{48} & \textbf{53}\\
    MemReasoner + 100\% SF & 56 & 44 & \underline{43} & \underline{38}\\
    \end{tabular}
    }
\end{table}

\begin{table}[t]
    \centering
    \caption{Performance on VT 2-hop → VT 1-hop generalization.}
    \label{tab:task2to1VT}
    \resizebox{0.8\columnwidth}{!}{
    \begin{tabular}{c|cccc}
    Model type & 0k & 1k & 2k & 4k\\\hline
    \textcolor{black}{RMT}& 0.8 & 0.6  & 0.2 & 0.1\\
    \textcolor{black}{RMT + 100\%SF}& 58.3 & 1.8 & 2.0 & 0.8 \\
    Mamba& 1.2 & 3.1 & 2.1 & 0.0 \\
    Mamba  + 100\%SF & 1.2 & 3.8 & 1.4 & 0.0 \\
    \hline
    MemReasoner & \textbf{100} & \textbf{100} & \textbf{100} & \textbf{100}\\
    \end{tabular}
    }
\end{table}

\subsection{Conclusion and Limitations}
Here we empirically investigate and compare generalization aspects of   long-range dependency modeling across a variety of LLM architectures, including RMT, Mamba, and the proposed MemReasoner. The MemReasoner model 
comes with an external-to-transformer latent memory of context, which aims to learn  
temporal relations and meaningful hopping between facts, and enables a selective attention over the memory. These sets of operations, when fine-tuned with supervision on the final answer alone, are found effective in generalization in single-hop scenarios, when compared to the internal (to transformer) segment-recurrence mechanism within RMT and the selection mechanism in Mamba. On more challenging two-hop tasks, MemReasoner utilizes additional supervision on intermediate reasoning steps more effectively than other models, and shows stronger and more robust generalization in most experiments. MemReasoner's performance benefits from having supporting facts for as little as 1\% training samples. The performance gain with supervision can depend on a number of factors (task, model, loss, testbed,  and  amount of supervision) which will be further explored in future work.  

The current work is limited to testing the MemReasoner framework on simple synthetic reasoning tasks, that only encompass very basic skills such as fact chaining and order awareness. Further, the datasets used are of simple construct. Future work will extend the framework to investigating generalization on more real-world reasoning tasks and datasets. 
and train the models in a more task-general setting. Another possible direction  to explore is defining a selective attention over a combination of  token-level memory and  segment-level memory. In general, designing alternative transformer architectures with new loss objectives that encourage the model to learn the underlying reasoning skills will a path to explore toward more robust reasoners.

\bibliography{ref}
\bibliographystyle{icml_styles/icml2025}


\newpage
\appendix
\onecolumn
\section*{Appendix}
\section{Algorithm}
\label{app:algorithm_description}
\begin{algorithm}[h]
\caption{}\label{alg:inference}

\DontPrintSemicolon
\SetKwFunction{FIterativeRead}{MemReasoner}
  \SetKwFunction{Ftempenc}{temporalEncoding}
  \SetKwFunction{Fenc}{encode}
  \SetKwFunction{Fdec}{decode}
  \SetKwFunction{Fwrite}{write}
  \SetKwFunction{Fread}{read}
  \SetKwFunction{Fqueryupdate}{queryUpdate}
  \SetKwFunction{Finference}{inference}
  \SetKwFunction{FIS}{InferenceTimeUpdate}
  \SetKwRepeat{Do}{do}{while}
  \SetKwProg{Fn}{Function}{:}{}
  \Fn{\FIterativeRead{~$q, \{c_1, ..., c_E\}, \alpha, \tau$}}{ 
    \tcp{$q$: query tokens,
    \newline $\{c_1, ..., c_E\}$: $E$ context lines, 
    \newline $\alpha$: a hyperparameter for the query update, 
    \newline $\tau$: a threshold hyperparameter for terminating iterations, 
    \newline $P_a$: the prompt given to the decoder for answer generation}
    $z_q \gets$ \Fenc{$q$} \\
  \For{$i \gets 1$ \KwTo $E$}{
        $z_i \gets$ \Fenc{$c_i$}
    } 
    $\tilde{z}_1, ..., \tilde{z}_E \gets$ \Ftempenc{$z_1, ..., z_E$} \;
    $\hat{M} \gets $\Fwrite{$\tilde{z}_1, ..., \tilde{z}_E$}  \;
    $\tilde{z}_r \gets$ \Fqueryupdate{$z_q, \alpha, \tau, \hat{M}$} \;
    (optional) $\tilde{z_r}, \{z_1,\ldots, z_E\}, \{\tilde{z}_1,\ldots, \tilde{z}_E\}\gets$ \FIS{$\tilde{z_r},z_1,\ldots, z_E,\tilde{z}_1,\ldots, \tilde{z}_E$}\;
    $i^* \gets \argmin_{i \in \{1 ... E\}} ||\tilde{z}_i - \tilde{z}_r||_2$ \;
        \KwRet \Fdec{$z_{i^*}$, $W_M$, $P_a$} \tcp*{generate the answer with the decoder, $W_M$ is a learnable parameter which interfaces the $z_{i^*}$ with the decoder}
  }
  \;


  \Fn{\Fqueryupdate{$z_q, \alpha, \tau, M$}}{
    $\tilde{z}_r \gets$ \Fread($W_qz_q, M$) \tcp*{$W_q$ is learned parameter}
    $z_q = z_q + \alpha \tilde{z}_r$ \tcp*{query update}
    $\tilde{z}_{r, \text{next}} \gets$ \Fread($W_qz_q, M$) \;
    \Do{$||\tilde{z}_{r, \text{next}} - \tilde{z}_r||_2 > \tau$}{ 
        $\tilde{z}_r \gets \tilde{z}_{r, \text{next}}$ \;
        $z_q = z_q + \gamma \tilde{z}_r$ \;
        $\tilde{z}_{r, \text{next}} \gets$ \Fread($W_qz_q, M$) \;
    }
        \KwRet $\tilde{z}_r$ \;
  }
  \;
\Fn{\FIS{$\tilde{z_r},z_1,\ldots, z_E,\tilde{z}_1,\ldots, \tilde{z}_E, \eta=125$}}{
Index set $\mathcal{I}\gets$ gather indices $i$, s.t. $\tilde{z}_i$ has the top $\eta$ smallest  distance to $\tilde{z}_r$ \tcp*{filter out the most irrelevant contexts}
$\{\tilde{z}_j\}_{j\in\mathcal{I}} \gets$ \Ftempenc{$\{z_j\}_{j\in\mathcal{I}}$} \;
$\hat{M} \gets $\Fwrite{$\{\tilde{z}_j\}_{j\in\mathcal{I}}$}  \;
$\tilde{z}_r \gets$ \Fqueryupdate{$z_q, \alpha, \tau, \hat{M}$} \;
\KwRet $\tilde{z_r}, \{z_j\}_{j\in\mathcal{I}}, \{\tilde{z}_j\}_{j\in\mathcal{I}}$
}
\end{algorithm}

\section{Data Details}
\label{app:dataset_details}
\begin{figure}
    \centering
    \includegraphics[width=0.9\linewidth]{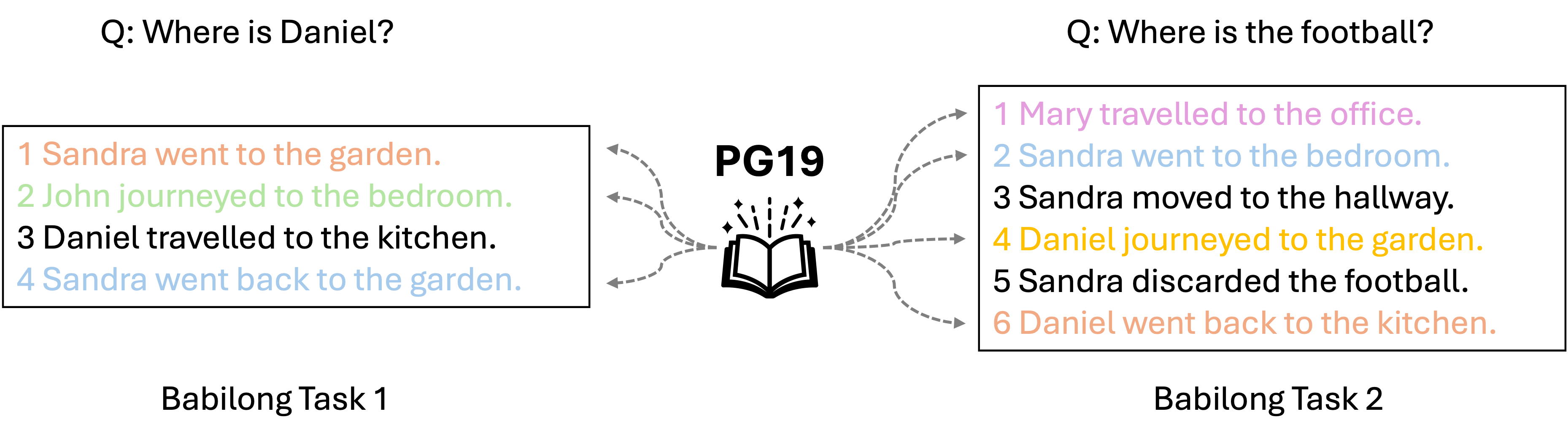}
    \caption{Examples of BABILong tasks.}
    \label{fig:example}
\end{figure}
\subsection{Additional dataset preprocessing details}
In the unprocessed bAbi data, a single data instance consists of a sequence of lines representing facts to reason over with questions interspersed throughout the facts.  We preprocess the bAbi data such that after pre-processing, a single training sample consists of a single question with facts for reasoning being the lines before it, with previous questions replaced by an empty line. On average, this leads to about 2 empty lines per training sample. For batches containing training samples with different lengths of context episodes, we pad shorter episodes with rows of the encoder padding token at the beginning.

During training, we also use Wikipedia samples which are chunked to a length of 64 tokens.

\subsection{Generation of location change dataset}
In order to test generalization to changes in location, we take each bAbi sample and map locations mentioned in the bAbi data to new out-of-distribution locations.  Specifically, bAbi data contains the mention of the following 6 locations: office, garden, hallway, bathroom, bedroom, and kitchen.  We replace all instances of these location names via the following mapping: office $\to$ library, garden $\to$ garage, kitchen $\to$ cafe, bathroom $\to$ attic, bedroom $\to$ basement, and hallway $\to$ gym.

\subsection{Generation of BABILong-soft data}
To generate BABILong-soft data, we begin with the bAbi test set for tasks 1 and 2.  We pad the data with sentences of the form:
\begin{verbatim}  {entity} {verb} to the {location}. \end{verbatim}
where the entity is one of ``John", ``Mary", ``Sandra", and ``Daniel", verb is one of ``moved", ``went", ``journeyed", ``travelled", ``went back", and location is one of ``bedroom", ``bathroom", ``kitchen", ``garden", ``office", and ``hallway". Sentences of this form naturally occur in the bAbi train and test set for these tasks.  The location of the padding is randomly sampled and thus this padding can occur before, between, and after the true sentences in the context of the bAbi dataset.  In order to avoid changing the answer to the bAbi question when adding padding, we keep track of the location of the last supporting fact and ensure that padding added after the last supporting fact does not mention the entity that is in the supporting fact.  Meanwhile, padding added before the last supporting fact can be have any entity.

To illustrate, consider a bAbi sample with the question ``Where is the football?" and context:
\begin{verbatim}
    Mary travelled to the office.
    Sandra went to the bedroom.
    Sandra moved to the hallway.
    Daniel journeyed to the garden.
    Sandra discarded the football.
    Daniel went back to the kitchen
\end{verbatim}
Here, the last supporting fact in the line ``Sandra moved to the hallway." and this line tells us the answer of the question is ``hallway".  To generate a corresponding BABILong-soft data point, we randomly sample a location at which we insert a sentence of padding.  If this line is to be added before ``Sandra moved to the hallway." then we sample any entity of the bAbi entities for the padding.  If it is to be added after ``Sandra moved to the hallway.", then we sample an entity that is not Sandra.

We use this approach to generate padding to increase context lengths to 1k, 2k, and 4k tokens.

\subsection{Additional details about Variable Tracking}\label{app:var_track_gen}
To generate variable tracking data for evaluation, we sample a letter of the alphabet and have that letter repeated 5 times as the variable name (ie. AAAAA).  Variable tracking data has 2 parameters, the number of hops and the number of chains.  The number of chains specifies the number of distinct numerical values present in the data.  For 1-hop data, each line of context has a variable assigned to each numerical value (ie. ``VAR AAAAA = 16438").  In 2-hop data, there are 2 variables assigned each value, but one of the variables references the other in assignment (ie. ``VAR AAAAA = 16438, VAR BBBBB = AAAAA").  Thus, the number of distinct variables is given by the number of hops times the number of chains.

We provide a data sample below with 2 hops and 2 chains with question `` Find all variables that are assigned the value 13075 in the text above":
\begin{verbatim}
    VAR DDDDD = 13075
    VAR FFFFF = 19367
    VAR ZZZZZ = VAR FFFFF
    VAR YYYYY = VAR DDDDD
\end{verbatim}

For RMT and Mamba we provide the context in the prompt so that the prompt looks as follows ``[INST] Memorize and track the chain(s) of variable assignment hidden in the following text. VAR DDDDD = 13075 VAR FFFFF = 19367 VAR ZZZZZ = VAR FFFFF VAR YYYYY = VAR DDDDD Question: Find all variables that are assigned the value 13075 in the text above. [/INST] Answer: According to the chain(s) of variable assignment in the text above, 2 variables are assigned the value 13075, they are:"

For MemReasoner, the query is taken to be ``[INST] Question: Find all variables that are assigned the value 13075 [/INST]" and the prompt we use for obtaining the answer is ``Answer: According to the chain(s) of variable assignment, 2 variables are assigned the value 13075, they are:"  During the training of MemReasoner, we also insert empty lines of context, with 0 empty lines added at chain length 2, 1 empty line added at chain length 4, 2 at chain length 6, 3 at chain length 8, and 4 and chain length 10 similar to how empty lines are present in training MemReasoner on bAbi.  The locations of these empty context lines are chosen randomly for variable tracking.

In order to increase the length of variable tracking data to 1k, 4k, and 16k token lengths, we follow the approach used by \citep{hsieh2024rulerwhatsrealcontext} in the RULER benchmark and pad by repeatedly inserting the noise: ``The grass is green. The sky is blue. The sun is yellow. Here we go. There and back again." until we reach the desired token length.  We note that this inserted noise can appear before, between, or after lines of variable tracking data.

\section{Training Details}
\label{sec:training}
We always initiate MemReasoner finetuning from Larimar checkpoint pretrained on Wikitext (obtained by following the training protocol described in \citep{das2024larimarlargelanguagemodels}), which uses a Bert-large as the encoder and a GPT2-large as the decoder. The number of parameters in MemReasoner reported in the main paper is 1.4B.  For extension to MemReasoner with a GPTJ-6B decoder, see section \ref{sec:gptj6b}. The slot size in the memory is 512.  
During finetuning, 
we randomly sample a batch of pretraining data (Wikipedia) of the same size as the batch of bAbi finetuning data, for computing the autoencoding loss on the pretrain dataset of 2M samples. 
 We generate the answer to the question by passing a prompt to the decoder (i.e. in the case of bAbi Task1-2, the prompt has the from ``$<$BOS$>$ $X$ is in the" where $X$ denotes subject of the query). 
 
\textit{bAbi training.} We train MemReasoner models for 200 epochs using Adam optimizer with learning rate 5e-6. We set batch size to be 10.  Additionally, we set query update parameter $\alpha=1$. The maximum episode length varies from 14 (bAbi Task 1) to 72 (bAbi Task 2). Which means that MemReasoner has been exposed to a maximum of 90 and 573 tokens during finetuning on task 1 and task 2, respectively, whereas at test-time the model is exposed to contexts that are up to 128k tokens long. Since bAbi Task 1 is a single hop task, we do not perform query update during either training or inference. When fine-tuning on bAbi Task 2, we perform a fix number of 2 hop (equivalent to 1 query update) during the training. With bAbi Task 2 fine-tuned MemReasoner, we re-use the ``2 hop'' setting at inference on all tasks, including bAbi Task 2 and BABILong Task1/2. We consistently use query update parameter $\alpha$=8 throughout our experiments and include an ablation study on $\alpha$ in the appendix. Due to the page limit, we also defer ablation studies on the episodic memory, temporal encoding schemes, use of inference time update, and the number of training epochs to Appendix \ref{app:ablation}. 

For RMT-0.77B and Mamba-1.4B finetuning, we finetune starting from off-the-shelf model checkpoints using the next token prediction loss on final answer reconstruction and optionally on supporting fact (SF) reconstruction (100\% SF) on each tasks separately  till the testing accuracy on the task is sufficiently high (near 100\%). 100\% SF models are trained to jointly reconstruct the answer given the prompt and output supporting facts when given a separate prompt asking for the supporting fact (i.e., when prompted ``Where is the apple? Supporting facts:", the model is trained to follow up with the text of each supporting fact). In practice, we train for 20 epochs with batch size 10 on Mamba for the accuracy to plateau. RMT training was done with multiple segments using a curriculum learning procedure. In order to train with more segments while exposing the model to only bAbi data, we reduce the segment size to 64 for task 1 and 128 for task 2.  This leads to 2 segments in training for task 1 and 2-4 segments in training for task 2.  In order to mimic the curriculum learning process, we filter the data so that we train with inputs with token length up to the segment size for 10 epochs with batch size 5, up to 2 times segment size for another 10 epochs, and so on. For both RMT and Mamba, we use Adam optimizer with learning rate $1e-5$.

\textit{Variable tracking (VT) training. }For the training on VT tasks, we do not use autoencoding loss on wiki samples. We train MemReasoner for 100 epochs. 
For VT, the episode length is at most 14 for both single hop and 2-hop tasks.  Similar to bAbi, we do not perform query update for the single hop task and perform a single query update for the 2-hop task. Since VT asks for all variables with a specific value, for MemReasoner, we take all unordered readouts of the model and pass them individually to the decoder to get the variables from each reasoning hop, and then concatenate these variables in order to obtain the final answer.

For RMT and Mamba on VT, we use the same experimental setup as with training on bAbi. Due to the shorter context lengths in variable tracking compared to bAbi, we train RMT with 2 segments, with segment size set to the median length on the train dataset. Similar to bAbi training, we use a curriculum schedule of training with 1 segment for 10 epochs and with 2 segments for an additional 10 epochs and train with a batch size of 5.

\section{Additional Experimental Results}
\label{app:additional_experimental_details}

\subsection{Comparison of inference-time complexity}
\label{app:complexity}
Let $H_1$, $H_2$ and $d_1$, $d_2$ be the number of transformer layers and hidden state dimension in the encoder and decoder, respectively. Let $E$ denote the number of context lines in a sample, $L$ be the max context length, $L_1$ be the max query length, $D$ be the latent space dimension, and $m$ be the memory size. The inference-time computational complexity for MemReasonr can be estimated by the encoder complexity $\mathcal{O}(H_1((EL^2+L_1^2)d_1+(EL+L_1)d_1^2))$, temporal encoding complexity $\mathcal{O}(Ed^2)$, memory operation complexity $\mathcal{O}(Edm^2)$, decoding complexity $\mathcal{O}(H_2(|P_a|^2d_2+|P_a|d_2^2))$, and broadcasting complexity $\mathcal{O}(d_1dE)$ and $\mathcal{O}(d_2dH_2)$. For a typical GPT decoding, the inference-time computational complexity is $\mathcal{O}(H_2((EL+L_1)^2d_2+(EL+L_1)d_2^2))$.

To provide a more direct comparison, we give in Table~\ref{tab:runtime_compare} the inference cost measured in seconds per input for evaluating with BABILong in comparison to the base decoder (gpt2-large). We note that gpt2-large does not support context lengths longer than 1024 tokens.  Overall, we observe that the increase in inference time for both RMT and MemReasoner are very small for 0k, and MemReasoner's runtime advantage over RMT becomes clear as they process longer context lengths. This is attributed to the utlization of latent encodings of contexts, performing one-shot write to the memory, and executing multiple hops over that memory, all in the latent space.
\begin{table}[h]
    \centering
    \caption{\textcolor{black}{The inference cost measured in seconds per input on BABILong.}}
    \label{tab:runtime_compare}
    \vspace{0.1in}
    \begin{tabular}{c|ccccccccc}
    Model type & 0k & 1k & 2k & 4k & 8k & 16k & 32k & 64k & 128k\\\hline
    gpt2-large & 0.28 & 1.13 & - &- &- &- &- &- &- \\
    RMT & 0.35 & 0.35 & 0.69 & 1.46 & 2.97 & 5.96 & 12.25 & 23.92 & 48.40\\
    MemReasoner     &  0.30 & 0.33 & 0.40 & 0.61 & 0.98 & 1.94 & 3.26 & 11.25 & 13.77
    \end{tabular}
    \vspace{-0.1in}
\end{table}
\subsection{Performance on original bAbi Test Set}
\label{app:babi}
Table~\ref{tab:babi} reports the performance of MemReasoner, RMT, and Mamba, each of which is independently finetuned  on original bAbi task 1  and task 2, on the corresponding bAbi test set of 1k samples. If supporting fact supervision is used during fine-tuning, it is mentioned explicitly as + x\% sup, where x denotes amount of training samples that have corresponding supporting facts. We will also include off-the-shelf powerful baselines    GPT-3 (175 B parameters)  with few-shot  and chain-of-thought prompting \citep{yang2023coupling}. Results show that, while prompting techniques such as few-shot learning and CoT prompting work well on task 1 which requires a single hop to find the entity location, those baselines perform much poorly on task 2 that requires learning temporal dependence  and performing multiple hops across facts to generate the final answer of object location. RMT and Mamba, when finetuned on bAbi, achieves near-perfect accuracy on both tasks. In contrast, MemReasoner trained with supervision on final answer, while perform comparably on  task 1, it falls short on task 2. This low test accuracy of MemReasoner is due to making frequent errors at the second hop. Thus, we add   supporting fact supervision on 1\% of training samples, which boosts the accuracy from 39.5\% to 74.5\%. When trained with supporting fact  supervision on all samples, the accuracy becomes near perfect. This result suggest MemReasoner requires a small amount of supporting fact supervision to perform two-hop reasoning in a reasonable manner. 
\begin{table}[t]
    \centering
    \caption{Performance on  original bAbi test set. Best model is highlighted in bold. GPT-3 (=text-davinci-003) baselines are from ~\citep{yang2023coupling}. Finetuning data, if any, seen by a model is specified within parentheses. }
    \label{tab:babi}
    \vspace{0.1in}
    \begin{tabular}{c|cc}
    Model type & Task 1 & Task 2\\\hline
    CoT - GPT-3 & 97.3 & 72.2\\
    Few-shot - GPT-3 & 98.4 & 60.8\\
    \hline
    \textcolor{black}{RMT-.77B (bAbi)} & 97.7 & 97.5\\
    Mamba-1.4B (bAbi) & \textbf{100} & 95\\
    \hline
    MemReasoner-1.4B (bAbi) & 100 & 39.5\\
    MemReasoner-1.4B (bAbi) + 1\% SF & - & 74.4\\
    MemReasoner-1.4B (bAbi) + 100\% SF & - & 99.6\\
    \end{tabular}
    \vspace{-0.1in}
\end{table}
\vspace{1em}

\subsection{Robustness of models finetuned on BABILong}
\citet{kuratov2024babilong} observe that directly finetuning with long contexts following the BABILong distribution leads to improved generalization to longer contexts.  We test the robustness of RMT-.14B finetuned on BABILong to unseen locations using our bAbi location change dataset (Table \ref{tab:BABILong_FT_location_swap}).  We observe that RMT finetuned on BABILong exhibits a large drop in accuracy on these unseen locations even in the absence of padding.
   \begin{table}[t]
    \centering
    \caption{RMT's robustness to location changes in bAbi test set when fine-tuned on BABILong. Mamba's result is unavailable due to the absence of publicly released finetuned checkpoint.}
    \label{tab:BABILong_FT_location_swap}
    \vspace{0.1in}
    \begin{tabular}{c|cc}
    Model type & Task 1 & Task 2\\\hline
    RMT-.14B (BABILong) & 12.4 & 1.6
    \end{tabular}
\end{table}

\subsection{\textcolor{black}{Extension to GPTJ-6B}}
\label{sec:gptj6b}
\textcolor{black}{MemReasoner is a model-agnostic way to augment current decoder-only LLMs with dynamically updatable memory. Via end-to-end training, the architecture learns to write the latent encodings in a fixed-size memory, order them in their order of appearance in the context, and perform multiple hop over that context and update the latent query accordingly. The decoder learns a differentiated attention mechanism to the readout from the memory,  to accurately generate the final answer and supporting facts (intermediate hops). Supervision on all supporting facts was enabled in the form of SF ordering loss and reconstruction loss.   Below, we provide the results when we train a GPTJ-6B decoder with MemReasoner training protocol, suggesting more or less similar performance compared to MemReasoner-1.3B.} 

\begin{table}[h]
    \centering
    \caption{\textcolor{black}{Performance of MemReasoner + 100\% SF  with a GPTJ-6B decoder on BABILong.}}
    \label{tab:gptj-6b}
    \vspace{0.1in}
    \begin{tabular}{c|ccccccccc}
    Model type & 0k & 1k & 2k & 4k & 8k & 16k & 32k & 64k & 128k\\\hline
    Task 1 & 98 & 82 & 77 & 65 & 60 & 68 & 70 & 65 & 67\\
    Task 2     &  98 & 65 & 50 & 34 & 35 & 32 & 22 & 27 & 30
    \end{tabular}
    \vspace{-0.1in}
\end{table}

\subsection{Ablation Studies}
\label{app:ablation}
\subsubsection{Inference Time Update}
We compare the performance of MemReasoner with and without inference time update (IU) as shown below:

\begin{table*}[h]
    \centering
    \caption{Ablation with and without IU for BABILong Task 1}
    \label{tab:BABILong1}
    \vspace{0.1in}
    \resizebox{0.82\textwidth}{!}{
    \begin{tabular}{cc|cccccccccccccccccccccc}
     & & Avg. & Avg.  & \\ \hline
    Model type & + IU & $\leq$ 8k & $\geq$ 16k & 0k & 1k & 2k & 4k & 8k & 16k & 32k & 64k & 128k \\

    \hline
    MemReasoner-1.4B (bAbi) &  & 77.2 & 62.8 & 97 & 82 & 73 & 65 & 69 & 62 & 66 & 63 & 60\\
   MemReasoner-1.4B (bAbi) & $\checkmark$ & 91.6 & 89 & 97 & 90 & 89 & 89 & \textbf{93} & \textbf{94} & \textbf{94} & 87 & 81\\\hline
    MemReasoner-1.4B (bAbi) + 100\% sup &  &82.2 & 59.3 & \textbf{100} & 91 & 80 & 70 & 70 & 62 & 58 & 58 & 59\\
   MemReasoner-1.4B (bAbi) + 100\% sup &  $\checkmark$ &\textbf{94.4} & \textbf{93} & \textbf{100} & \textbf{93} & \textbf{93} & \textbf{94} & 92& \textbf{94} & \textbf{94} & \textbf{94} & \textbf{90}\\

    \end{tabular}}
\vspace{1em}
    \centering
    \caption{Ablation with and without IU for BabiLong Task 2}
    \label{tab:BABILong2}
    \resizebox{0.82\textwidth}{!}{
    \begin{tabular}{cc|ccccccccccccccccccccc}
   &    & Avg. & Avg.  & \\ \hline
    Model type & + IU & $\leq$ 8k & $\geq$ 16k & 0k & 1k & 2k & 4k & 8k & 16k & 32k & 64k & 128k \\\hline
    MemReasoner-1.4B (bAbi) &  & 18.4 & 11.8 & 38 & 16 & 12 & 14 & 12 & 8 & 14 & 14 & 11\\
    MemReasoner-1.4B (bAbi) & $\checkmark$ & 21 & 18.8 & 
    38 & 17 & 17 & 16 & 17 & 15 & 15 & 27 & 18\\ \hline
    MemReasoner-1.4B (bAbi) + 1\% sup & & 42.8 & 27.5 & 76 & 47 & 33 & 31 & 27 & 25 & 24 & 31 & 30\\
     MemReasoner-1.4B (bAbi) + 1\% sup & $\checkmark$ & 46.6 & 27.8 & 76 & 43 & 46 & 38 & 30 & 27 & 25 & 31 & 28\\ \hline
    MemReasoner-1.4B (bAbi) + 100\% sup &  & 59.4 & 24.5 & \textbf{100} & \textbf{70} & 52 & 44 & 31 & 28 & 20 & 24 & 26\\
    MemReasoner-1.4B (bAbi) + 100\% sup & $\checkmark$ &\textbf{74.2} & \textbf{59.5} & \textbf{100} & 64 & \textbf{71} & \textbf{63} & \textbf{73} & \textbf{71} & \textbf{61} & \textbf{55} & \textbf{51}\\ 
    \end{tabular}}
    \vspace{-0.1in}
\end{table*}

\subsubsection{Memory}
In Table~\ref{tab:wogpm}, we conduct the ablation study on the episodic memory module in MemReasoner on bAbi and BABILong, task 1 and 2. Specifically, MemReasoner w/o memory module uses the same architecture of encoder and decoder (BERT-Large and GPT2-Large respectively) but does not use the memory module for encoding the context.  Instead, the MemReasoner w/o memory uses the encoder to encode only the question and this is passed in to the decoder as kv-cache.  Additionally, the context and question are passed to the decoder as part of the prompt with the format:

\begin{verbatim}
Context:\n{context}\nQuestion:\n{question}\nAnswer: 
\end{verbatim}
where \{context\} and \{question\} represent the context and the question for the datapoint.  We train the model with reconstruction loss to ensure that the model is able to fill in the answer given this prompt and with autoencoding loss on the pretraining dataset (see last term of Equation \ref{train_loss}) in order to reduce overfitting on bAbi data.  We train MemReasoner w/o memory module for 5 epochs.

MemReasoner w/o memory module trained on bAbi task 1 obtains almost perfect accuracy on bAbi task 1 and BABILong task 1 0k. However, its generalization ability to long context (BABILong 1k and 2k) is much inferior to MemReasoner (MemReasoner$\backslash$memory 0\% vs. MemReasoner 91\% on BABILong 1k). Similar trends can also be seen from bAbi task 2 trained MemReasoner$\backslash$memory, implying the significance of the episodic memory module and the operations around it in MemReasoner. 
\begin{table}[h]
    \centering
    \caption{Ablation study on the episodic memory}
    \label{tab:wogpm}
    \vspace{0.1in}
    \begin{tabular}{c|cccc|cccc}
    Model type     &  Task 1 & 0k & 1k & 2k & Task 2 & 0k & 1k & 2k\\\hline
    MemReasoner$\backslash$memory & \textbf{100}  & \textbf{100} & 0& - & 99.3 & \textbf{100} & 29 & -\\
    MemReasoner & \textbf{100} & 99 & \textbf{91} & \textbf{83} & \textbf{100} & \textbf{100} & \textbf{73} &\textbf{ 61}
    \end{tabular}
    \vspace{-0.1in}
\end{table}

\subsubsection{Temporal Encoding}
\label{app:temporal}
The temporal encoding module is generic and allows any structure featuring sequentiality within context. In the paper, we have used \textit{parameterized} methods such as GRUs. In practice, we can also use \textit{un-parameterized} methods such as Sinusoidal Positional Encoding~\citep{vaswani2017attention}.
Additionally, we experiment with positional encoding which assigns encodings starting from the last element of the episode.
This structure ensures that for contexts of different length, the last lines of the contexts are encoded similarly, which is useful for QA tasks in which the most recent information is more relevant for answering the question.
Finally, to convert $\{z_1, ..., z_E\}$ to $\{\tilde{z}_1, ..., \tilde{z}_E\}$ with positional encodings, we add the computed positional encodings to the input.

In Table~\ref{tab:position_scheme}, we experiment with different temporal encoding schemes, including non-parametric method (Positional Encoding) and parametric method (GRU). In the table, we show MemReasoner's accuracy on BABILong Task 1. It can be seen that GRU encoding has significant advantage over Positional Encoding, with much slower decay in the accuracy as the context length increases. Additionally, though showing higher accuracy compared with Positional Encoding, uni-directional GRU's accuracy decreases faster than bi-directional GRUs. Since 1-layer bi-directional GRU has similar performance with 2-layer bi-directional GRU, we choose the lighter model and use 1-layer bi-directional GRU throughout the experiments in this paper.
\begin{table}[h]
    \centering
    \caption{Ablation study on the temporal encoding schemes.}
    \label{tab:position_scheme}
    \vspace{0.1in}
    \begin{tabular}{c|ccccccccccccccc}
    Encoding scheme &  0k & 1k & 2k  \\\hline
    Positional Encoding     & \textbf{100} & 27 & 20\\
    2-layer bi-directional GRU & \textbf{100} & 90 & 80\\
    2-layer uni-directional GRU & 94 & 75 & 61\\
    1-layer bi-directional GRU & 99 & \textbf{91} & \textbf{83}
    \end{tabular}
    \vspace{-0.1in}
\end{table}

\subsubsection{Query Update $\alpha$}
In Table~\ref{tab:inference_hprm}, we exploit test-time inference hyper-parameter $\alpha$ and its effect in reasoning tasks' performance. We draw inspiration from ~\citep{kollias2024generation}, where authors investigated the effect of scaling readout vectors to improve generation quality. In Line 20 of Algorithm~\ref{alg:inference}, when using an $\alpha>1$, we equivalently scale up the readout vectors which greatly help our generalization to Task 1 BABILong according to Table~\ref{tab:inference_hprm} (e.g. from 14\% to 45\% on 4k context token task).
\begin{table}[h]
    \centering
    \caption{Ablation study on the query update parameter $\alpha$.}
    \label{tab:inference_hprm}
    \vspace{0.1in}
    \resizebox{0.82\textwidth}{!}{
    \begin{tabular}{c|c|ccccccccc|ccccc}
   Query update & Task 2 bAbi & \multicolumn{9}{c|}{Task 2 BABILong} & \multicolumn{4}{c}{Task 1 BABILong}\\
    $\alpha$ & location change & 0k & 1k & 2k & 4k & 8k & 16k & 32k & 64k & 128k & 0k & 1k & 2k & 4k\\\hline
    1& 52.6 &  \textbf{100} & 46 & 25 & 18 & 18 & 13 & 16 & 12 & 13 & 78 & 21 & 17 & 14\\
    4& \textbf{54.2} &  \textbf{100} & \textbf{73} & \textbf{61} & \textbf{46} & \textbf{26} & \textbf{22} & \textbf{19} & \textbf{19} & \textbf{27} & \textbf{83} & 47 & 44 & 40\\
    8& 52.7 &   \textbf{100} & \textbf{73} & \textbf{61} & \textbf{46} & 23 & 20 & \textbf{19} & 17 & 20 & \textbf{83} & \textbf{58} & \textbf{50} & \textbf{45}
    \end{tabular}}
    \vspace{-0.1in}
\end{table}

\subsubsection{Training epochs}
In Table~\ref{tab:training_epochs}, we evaluate MemReasoner's performance when fine-tuned on bAbi task 2 as a function of the number of training epochs. Specifically, with fewer epochs, MemReasoner demonstrates stronger robustness to location change, reaching an accuracy of 79\% at the 66th epoch, which decreases to around 50\% as the training continues (at 100/200th epoch). On the other side, MemReasoner's accuracy on shorter context tasks in BABILong Task 1 and 2 (i.e. 0-4k) improves as the training continues.
\begin{table}[h]
    \centering
    \caption{Ablation study on the number of training epochs}
    \label{tab:training_epochs}
    \vspace{0.1in}
    \resizebox{0.82\textwidth}{!}{
    \begin{tabular}{c|c|ccccccccc|ccccc}
     & Task 2 bAbi & \multicolumn{9}{c|}{Task 2 BABILong} & \multicolumn{4}{c}{Task 1 BABILong}\\
    \#epochs & location change & 0k & 1k & 2k & 4k & 8k & 16k & 32k & 64k & 128k & 0k & 1k & 2k & 4k\\\hline
    66 & \textbf{78.0} & 99&70&54&30&27&23&17&\textbf{18}&17&58&51&45&37\\
    100 & 47.3 & \textbf{100} & 70 & 57 & 38 & \textbf{28} & \textbf{31} & \textbf{25} & 12 & 19 & 82 & \textbf{58} & \textbf{50} & \textbf{46}\\
    200 & 52.7 &  \textbf{ 100} & \textbf{73} & \textbf{61} & \textbf{46} & 23 & 20 & 19 & 17 & \textbf{20} & \textbf{83} & \textbf{58} & \textbf{50} & 45
    \end{tabular}}
    \vspace{-0.1in}
\end{table}

\subsection{MemReasoner with even stronger supporting fact supervision}
\label{app:support}
To further boost the expandability of a decoding result, one can optionally optimize the reconstruction loss of the supporting facts using the available supervised data. Specifically, if we let $P_s$ denote the prompts for generating the supporting fact, we can instead minimize
\tikzset{
h1/.style={
set fill color=blue!10,
set border color=blue!10,
},
h2/.style={
set fill color=red!10,
set border color=red!10,
}
}

\begin{equation}
\label{train_loss}
\begin{split}
 &L = \rho \underbrace{\mathbb{E}_{x \sim \mathcal{D}_{\text{pretrain}}} \ln p(d(e(x)))}_{\text{autoencoding of pretraining dataset}}\nonumber +\mathbb{E}_{(q, C, S, a) \sim \mathcal{D}_{\text{finetune}}}\underbrace{\mathbb{E}_{z_{r}^{|S|} \sim p(z_{r}^{|S|} | q, M, \tilde{z}_{r}^{0} ... \tilde{z}_{r}^{|S|-1})} \ln p(a | z_{r}^{|S|}, P_a)}_{\text{reconstruction of answer}
}  \nonumber\\
&\tikzmarkin[h2]{b}(2.45,-1.2)(-0.2,1.2) +\mathbb{E}_{(q, C, S, a) \sim \mathcal{D}_{\text{finetune}}^*} \left[ \delta \underbrace{\sum_{i=1}^{|S|} \mathbb{E}_{\tilde{z}_{r}^{i} \sim p(\tilde{z}_{r}^{i} |q, M, \tilde{z}_{r}^{0} ... \tilde{z}_{r}^{i-1})} \ell_{\text{order}}(\tilde{z}_{r}^{i}, S_i)}_{\text{ordering loss}} \right.\tikzmarkend{b}\nonumber\\
&\tikzmarkin[h1]{a}(0.25,-0.8)(-0.2,0.9)\left. +\alpha \sum_{i=1}^{|S|} \underbrace{\mathbb{E}_{z_{r}^{i} \sim p(z_{r}^{i} | M, \tilde{z}_{r}^{0} ... \tilde{z}_{r}^{i-1}) } \ln p(c_{S_i} | z_r^i, P_s)}_{\text{reconstruction of supporting facts}}  + \beta \sum_{s \in S} \underbrace{\ln p(d(e(c_s)))}_{\text{autoencoding of supporting fact}} \right],\tikzmarkend{a}\\
\end{split}
\end{equation}
where $\alpha$ and $\beta$ are hyperparameters controlling regularization strength of the two additional terms - 
the 4th term corresponds to the reconstruction loss of the supporting fact(s) with respect to the corresponding prompt for obtaining the answer $P_a$ and readout; the 5th term is the autoencoding loss of the supporting fact(s). We denote this comprehensive use of supporting fact as ``SF*''. MemReasoners trained with additional losses perform similarly on most task except transfer task generalization as shown in Table~\ref{tab:task2to1long_recon}. It should be mentioned that training RMT or Mamba with the ordering loss is non-trivial and is beyond the scope of current work.
\begin{table}[h]
    \centering
    \caption{Performance on bAbi task 2 → BABILong task 1 generalization.}
    \label{tab:task2to1long_recon}
    \vspace{0.1in}
    \begin{tabular}{c|cccc}
    Model type & 0k & 1k & 2k & 4k\\\hline
    MemReasoner-1.4B (bAbi) + 100\% SF & 56 & 44 & 43 & 38\\
    MemReasoner-1.4B (bAbi) + 100\% SF* & \textbf{83} & \textbf{58} & \textbf{50} & \textbf{45}
    \end{tabular}
    \vspace{-0.1in}
\end{table}

\subsection{Additional baselines}
\label{app:baselines}
We also add a Larimar-1.3B baseline, which is finetuned 
on Wikipedia and either bAbi or VT samples (indicated in parentheses for each table below),
with final answer reconstruction loss and autoencoding loss. The purpose of comparing MemReasoner with respect to Larimar is to disambiguate the benefits of temporal feature learning and iterative query and read updates on top of the episodic memory. Larimar fine-tuning shares the same training setups as MemReasoner. 

\begin{table*}[!h]
\centering
\caption{BABILong  Results. 
}
\label{tab:BABILong1-larimar}
\vspace{0.1in}
\begin{tabular}{c|cccccccccccccccccccccc}
& Avg. & Avg.  & \\
Model type & $\leq$ 8k & $\geq$ 16k & 0k & 1k & 2k & 4k & 8k & 16k & 32k & 64k & 128k \\\hline
    Larimar-1.3B (bAbi task 1) & 44.8 & 14.3 & 63 & 59 & 55 & 28 & 19 & 14 & 16 & 13 & 14\\
    \hline
    Larimar-1.3B (bAbi task 2)& 31 & 20.3& 42 & 41 & 29 & 22 & 21 & 19 & 16 & 22 & 24\\
 \end{tabular}
 \vspace{-0.1in}
 \end{table*}

\begin{table*}[!h]
\centering
\caption{Variable tracking results.
}
\label{tab:vt_single_hop-larimar}
\vspace{0.1in}
\begin{tabular}{c|cccc}
Model type & 0k & 1k & 4k & 16k \\
\hline
Larimar-1.3B (VT task 1) & 92.5 & 92.5 & 94.0 & 93.6 \\
 \hline
Larimar-1.3B (VT task 2) & 0.1 & 0 & 0.1 & 0 \\
\end{tabular}
\end{table*}

\begin{table}[!h]
    \centering
    \caption{Robustness to location changes in bAbi test set.
    }
    \label{tab:location_swap-larimar}
    \vspace{0.1in}
    \begin{tabular}{c|cc}
    Model type & Task 1 & Task 2\\\hline
    Larimar-1.3B (bAbi)& 24.9 & 7.8\\
    \end{tabular}
    \vspace{-0.1in}
\end{table}

\begin{table}[!h]
    \centering
    \caption{Performance on bAbi/VT task 2 →  task 1 (long) generalization.
    }
    \label{tab:task2to1long-larimar}
    \vspace{0.1in}
    \begin{tabular}{c|cccc}
    Model type & 0k & 1k & 2k & 4k\\\hline
    Larimar-1.3B (bAbi)& 45 & 19 & 20 & 11\\
    \hline
     Larimar-1.3B (VT)& 0.2 & 0.1 & 0.3 & 0.1\\
    \end{tabular}
    \vspace{-0.1in}
\end{table}

\end{document}